\documentclass[10pt,twocolumn,letterpaper]{article}

\usepackage{cvpr}              %

\usepackage{booktabs}
\usepackage[accsupp]{axessibility} %

\usepackage{times}
\usepackage{epsfig}
\usepackage{graphicx}
\usepackage{amsmath}
\usepackage{amssymb}
\usepackage{bbm}
\usepackage{pifont}
\usepackage{subcaption}
\usepackage{mathtools}
\usepackage{multirow}
\usepackage{adjustbox}
\usepackage[utf8]{inputenc}

\usepackage[dvipsnames]{xcolor}
\usepackage{xspace}

\usepackage[pagebackref,breaklinks,colorlinks,citecolor=LimeGreen]{hyperref}

\usepackage[capitalize]{cleveref}
\crefname{section}{Sec.}{Secs.}
\Crefname{section}{Section}{Sections}
\Crefname{table}{Table}{Tables}
\crefname{table}{Tab.}{Tabs.}

\newcommand{\wils}{WILSS}
\newcommand{\wilson}{WILSON}

\DeclareMathOperator*{\argmax}{arg\,max} %
\newcommand{\real}{{\rm I\!R}}
\newcommand{\myparagraph}[1]{\vspace{4pt}\noindent\textbf{#1}}
\newcommand{\set}{\mathcal}

\newcommand*\samethanks[1][\value{footnote}]{\footnotemark[#1]}

\def\cvprPaperID{10756} %
\def\confName{CVPR}
\def\confYear{2022}

\begin{document}

\title{\vspace{-10pt}Incremental Learning in Semantic Segmentation from Image Labels}

\author{
Fabio Cermelli\thanks{Equal contribution}\ $^{1,2}$, Dario Fontanel\samethanks\ $^{1}$, Antonio Tavera\samethanks\ $^{1}$, Marco Ciccone$^{1}$, Barbara Caputo$^{1}$\\
$^1$Politecnico di Torino, $^2$Italian Institute of Technology\\
{\tt\small \{first.last\}@polito.it} \\
}

\maketitle

\begin{abstract}
Although existing semantic segmentation approaches achieve impressive results, they still struggle to update their models incrementally as new categories are uncovered. Furthermore, pixel-by-pixel annotations are expensive and time-consuming. This paper proposes a novel framework for Weakly Incremental Learning for Semantic Segmentation, that aims at learning to segment new classes from cheap and largely available image-level labels.
As opposed to existing approaches, that need to generate pseudo-labels offline, we use a localizer, trained with image-level labels and regularized by the segmentation model, to obtain pseudo-supervision online and update the model incrementally.
We cope with the inherent noise in the process by using soft-labels generated by the localizer. We demonstrate the effectiveness of our approach on the Pascal VOC and COCO datasets, outperforming offline weakly-supervised methods and obtaining results comparable with incremental learning methods with full supervision.~\footnote{Code can be found at \href{https://github.com/fcdl94/WILSON}{https://github.com/fcdl94/WILSON}.}
\end{abstract}

\section{Introduction}
\label{sec:introduction}
Semantic segmentation is a fundamental problem in computer vision where significant progress has been made thanks to the surge of deep learning \cite{chen2017deeplab, chen2017rethinking, chen2018encoder} and the availability of large-scale human-annotated or synthetic datasets~\cite{Everingham2009ThePV, lin2014microsoft, Cordts2016Cityscapes, alberti2020idda, Richter_2016_ECCV}.
Despite the fact that many pre-trained models using public datasets are available online, one of their key disadvantages is that they are not meant to be incrementally updated over time and their knowledge is often limited to the predefined set of classes.

\begin{figure}[t]
    \centering
    \includegraphics[width=\linewidth]{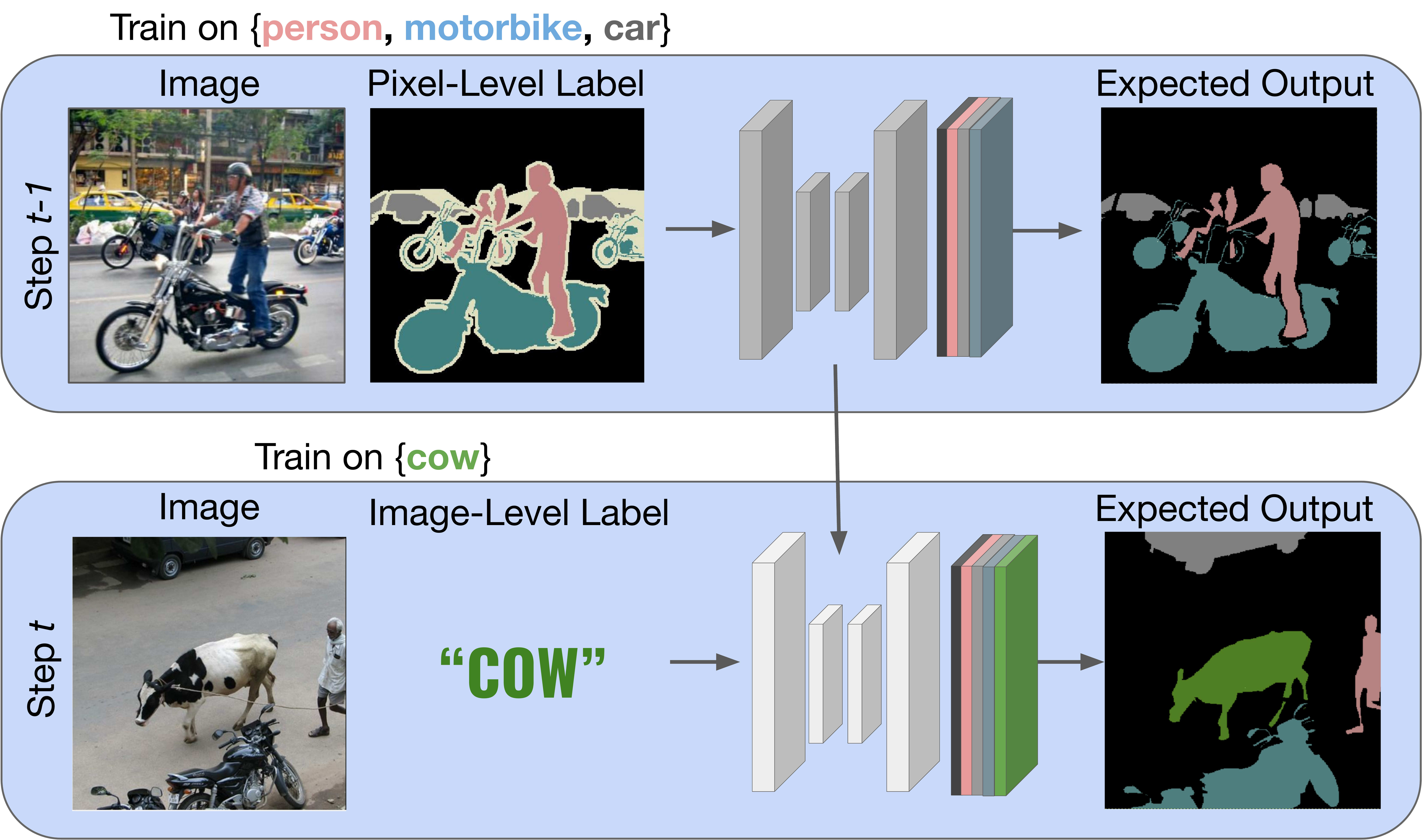}
    \vspace{-10pt}
    \caption{Illustration of \wils. A model is first pre-trained on a set of classes (\eg, \textit{person, motorbike, car}), using expensive pixel-wise annotations. Then, in the following incremental learning steps, the model is updated to segment new classes (\eg, \textit{cow}) being provided image-level labels and without access to old data.}
    \label{fig:teaser}
    \vspace{-10pt}
\end{figure}

A na\"ive solution to this problem would be to extend existing datasets with new annotated samples and train new models from scratch. 
However, this approach is impractical in case of frequent updates because training on the entire augmented dataset would take too long, increasing the energy consumption and carbon footprint of machine learning models~\cite{thompson2020computational, strubell2019energy, patterson2021carbon}.
Moreover, retraining or fine-tuning becomes infeasible when the original data is no longer available, \eg, due to privacy concerns or intellectual property.

A better solution is to incrementally add new classes to the pre-existing model, as done in some recent works~\cite{michieli2019incremental, cermelli2020modeling, douillard2021plop, michieli2021continual, maracani2021recall}. 
Incremental learning approaches update the model's parameters by training only on new data and employing ad-hoc techniques to avoid catastrophic forgetting on old classes~\cite{mccloskey1989catastrophic}.
While they reduce the cost of training, they rely on pixel-wise supervision on novel classes,  which is expensive and time-consuming to collect, and usually requires expert human annotators \cite{lin2014microsoft, bearman2016s}.

To reduce the annotation cost, different types of weak supervision have been proposed: bounding boxes~\cite{dai2015boxsup, khoreva2017simple}, scribbles~\cite{vernaza2017learning, lin2016scribblesup}, points~\cite{cheng2021pointly}, and image-level labels~\cite{pinheiro2015image, pathak2015constrained, kolesnikov2016seed}. %
Image labels can be easily retrieved from image classification benchmarks~\cite{deng2009imagenet} or the web, dramatically lowering the annotation cost. Nevertheless, their use has never been investigated in an incremental learning setting.

In light of these considerations, we argue that it is crucial to jointly address the problems of incrementally updating the model and reducing the annotation cost of new data for semantic segmentation. To this end, we propose to incrementally train a segmentation model using only image-level labels for the new classes. We call this task \textit{Weakly-Supervised Incremental Learning for Semantic Segmentation} (\wils). This novel setting combines the advantages of incremental learning (training only on new class data) and weak supervision (cheap and largely available annotations). An illustration of \wils\ is reported in \cref{fig:teaser}.

Directly applying existing weakly-supervised methods to incremental segmentation would require to (i) extract pixel-wise pseudo-supervision offline using a weakly supervised approach \cite{araslanov2020single, wang2020self, ahn2018learning, sun2020mining, lee2021railroad} and (ii) update the segmentation network resorting to an incremental learning technique \cite{cermelli2020modeling, douillard2021plop, maracani2021recall}.
However, we argue that generating pseudo-labels offline in incremental settings is sub-optimal, as it involves two separate training stages and ignores the model's knowledge on previous classes that can be exploited to learn new classes more efficiently.

Hence, we propose a \textbf{W}eakly \textbf{I}ncremental \textbf{L}earning framework for semantic \textbf{S}egmentation that incrementally trains a segmentation model generating \textbf{ON}line pseudo-supervision from image-level annotations (WILSON) and exploits previous knowledge to learn 
new classes.
We extend a standard encoder-decoder segmentation architecture~\cite{chen2017deeplab, chen2017rethinking, chen2018encoder} by introducing a \textit{localizer} on the encoder, from which we extract pseudo-supervision for the segmentation backbone. 
To improve the pseudo-supervision, we train the localizer with a pixel-wise loss
guided by the predictions of the segmentation model.
This regularization serves two purposes: i) it acts as a strong prior for the previous class distribution, informing the model on where old classes are located in the image, and (ii) it provides a saliency prior for extracting better object boundaries. 
To address the noise present in the pseudo-supervision,
instead of using hard pseudo-labels as in previous works
\cite{wang2020self, araslanov2020single, lee2021railroad}, we obtain soft-labels from the localizer, which provides information on the probability assigned to a pixel to belong to a certain class.

To summarize, the contributions are as follows:
\begin{itemize}
\setlength\itemsep{-0.4em}
    \item We propose the Weakly supervised Incremental Learning for Semantic Segmentation (\wils) task to extend pre-trained segmentation models with new classes using image-level supervision only.
    \item We propose \wilson, a novel framework that generates pseudo-supervision online using a simple localizer trained with an image-level classification loss and a pixel-wise localization loss that relies on old class knowledge. 
    To model the noise in the pseudo-supervision, we use a convex combination of soft and hard labels that improves the segmentation performance over hard labels only.
    \item We evaluate our method on the Pascal VOC~\cite{Everingham2009ThePV} and COCO~\cite{lin2014microsoft} datasets, showing that our approach outperforms offline weakly-supervised methods, 
    and that it is comparable or slightly inferior \wrt fully supervised incremental learning methods. 
    
\end{itemize}

\begin{figure*}[th]
    \centering
    \vspace{-10pt}
    \includegraphics[width=0.94\textwidth]{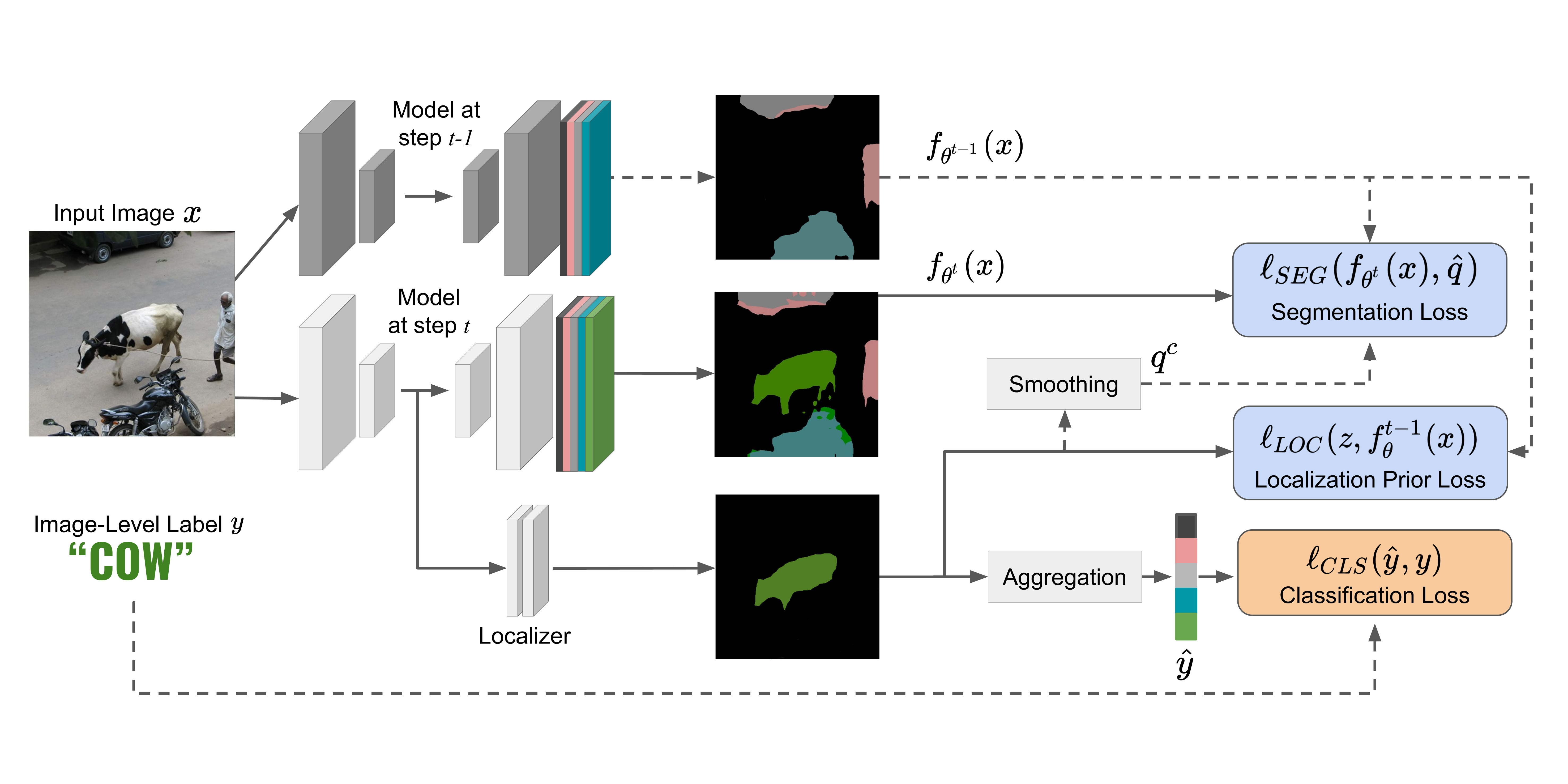}
    \caption{Illustration of the end-to-end training of \wilson. The localizer is directly trained using a classification loss $\ell_{CLS}$ and the Localization Prior loss $\ell_{LOC}$, which exploits the prior information of the old model at step $t-1$. The segmentation model is supervised using CAM and old model output. The gradient is not backpropagated on dotted lines.}
    \label{fig:method}
    \vspace{-10pt}
\end{figure*}

\section{Related work}
\label{sec:related}
\myparagraph{Incremental learning semantic segmentation.} Incremental learning (IL) aims at addressing the phenomenon known as \textit{catastrophic forgetting}~\cite{french1999catastrophic, mccloskey1989catastrophic}: a model, expanding its knowledge with new classes over time, gradually forgets previously learned ones. Even if in image classification it has been exhaustively studied \cite{rebuffi2017icarl, hu2021distilling, Zhu_2021_CVPR, ahn2021ss, fini2020online, liu2020mnemonics, li2017learning, dhar2019learning, chaudhry2018riemannian, shin2017continual, douillard2020podnet}, in semantic segmentation it is still in its early stages \cite{michieli2019incremental, michieli2021knowledge, cermelli2020modeling, douillard2021plop, michieli2021continual, klingner2020class, maracani2021recall, cermelli2020few}. 
As first shown by~\cite{cermelli2020modeling}, catastrophic forgetting in segmentation is exacerbated by the background shift problem; hence, they proposed a modified version of the traditional cross-entropy to propagate only the probability of old classes through the incremental steps and a distillation term to preserve previous knowledge. Later, \cite{douillard2021plop} proposed to preserve long and short-range spatial relationships at feature level, while \cite{michieli2021continual} regularized the latent space to improve class-conditional features separations. 
Alternatively, \cite{maracani2021recall} used samples of old classes with replay methods to mitigate forgetting. Finally, \cite{cermelli2020few} proposed the incremental few-shot segmentation setting, where only a few images to learn new classes are provided.

Differently from these works, we focus on a more challenging scenario where the supervision on new classes is provided as cheap image-level labels. 

\myparagraph{Weakly supervised semantic segmentation.} 
Collecting accurate pixel-wise annotations for supervising semantic segmentation models is generally costly and time-consuming. To address this issue, Weakly Supervised Semantic Segmentation (WSSS) methods aim to obtain effective segmentation models using cheaper supervisions such as bounding boxes~\cite{dai2015boxsup, papandreou2015weakly, khoreva2017simple}, scribbles~\cite{tang2018normalized, lin2016scribblesup}, points~\cite{bearman2016s,qian2019weakly}, and image-level labels~\cite{lee2019ficklenet, huang2018weakly, kolesnikov2016seed, sun2020mining}. 
Because of low prices and large availability on the web, image-level supervision gained the most attention over other types of weak supervision.
Most image-based weakly supervised approaches~\cite{kolesnikov2016seed, oh2017exploiting, huang2018weakly, ahn2018learning, ahn2019weakly, lee2019ficklenet, sun2020mining, chang2020weakly} use a two-stage procedure: (i) they generate pixel-wise pseudo-labels and then (ii) use them for training a segmentation backbone. The pseudo-labels are often extracted from an image-level classifier exploiting its Class Activation Maps (CAMs) \cite{zhou2016learning}. An exception is \cite{araslanov2020single}, which proposes to learn a segmentation model in a single stage.
Previous works focused on improving the pseudo-labels through multiple refinements steps~\cite{ahn2018learning, ahn2019weakly}, additional losses~\cite{kolesnikov2016seed, huang2018weakly, chang2020weakly, wang2020self, sun2020mining, araslanov2020single}, or erasing techniques that force the CAM to expand and focus on non-discriminative parts of the image ~\cite{wei2017object, hou2018self, chaudhry2017discovering}.
Finally, a recent trend uses external information, such as saliency, to improve the object boundaries \cite{lee2021railroad, yao2021non}.

Despite the rapid development of pseudo-labels generation techniques from image-level supervision, these works operate in a static scenario where the model learns from a fixed set of classes. Instead, we focus on the more challenging incremental learning setting where we learn new classes over time, extending a pre-trained segmentation model using only image-level labels.

\section{\wilson\ Framework} \label{sec:method}
Adapting current WSSS methods~\cite{wang2020self, lee2021railroad, li2021pseudo, ahn2018learning, kolesnikov2016seed} for incremental learning requires generating pseudo-labels offline for the new classes and then training a segmentation model separately. Instead, we propose an end-to-end framework for \wils\ that can learn incrementally from pseudo-labels generated online by a localizer attached to the model.
In the following, we first define the problem and the notation (Sec.~\ref{sec:problem}). Then we illustrate how the classification module can be trained to obtain pseudo-supervision (Sec.~\ref{sec:classifier}). Finally, in Sec.~\ref{sec:segmentation} we describe how to train the segmentation model to learn new classes without forgetting old ones. The framework is depicted in Fig.~\ref{fig:method}.

\subsection{Problem Definition and Notation} \label{sec:problem}
We consider an input space $\mathcal{X}$ (\ie the image space) and assume, without loss of generality, that each image is composed by a set of pixels $\set I$ with constant cardinality $|\set I| = H \times W = N$.
The output space $\mathcal{Y}^{N}$ is defined as the product set of $N$-tuples with elements in a label space $\mathcal{Y}$. 
In the standard semantic segmentation setting, given an image $x \in \mathcal{X}$, we want to learn a mapping to assign each pixel $x_i$ a label $y_i \in \mathcal{Y}$, representing its semantic class.  
The mapping is realized by a model $f_{\theta} = d_{\theta^d} \circ e_{\theta^e} : \mathcal{X} \mapsto \mathcal{\real}^{N \times |\mathcal{Y}|}$ from the image space $\mathcal{X}$ to a pixel-wise class probability vector. $e$ and $d$ denote the encoder and decoder of the segmentation network,  respectively.

The output segmentation mask is obtained as $y^* = \{ \argmax_{c\in\mathcal{Y}} p_i^c\}_{i=1}^{N}$, where $p_i^c$ is the model prediction of pixel $i$ for class $c$.

In the incremental segmentation setting~\cite{cermelli2020modeling}, training is realized over multiple \textit{learning steps}. At each learning step $t$, the previous label set $\set Y^{t-1}$ is augmented with novel classes  $\set C^t$, yielding a new label set $\set Y^t=\set Y^{t-1}\cup\set C^{t}$.
Differently from the original incremental setting, in {\wils} we are provided with dense annotations only for the initial step ($t=0$). That is, the model is pre-trained on a densely-annotated dataset $\mathcal{T}^0 \subset \mathcal{X}\times (\set {C}^{0})^N$ only for the initial classes. Then, we learn new classes only from cheap image-level labels for all the following steps. Namely, for ($t>0$), we have access to multiple training sets with only image-level annotations for novel classes $\mathcal{T}^t \subset \mathcal{X}\times (\set{C}^{t})$.
As in \cite{cermelli2020modeling}, we assume that data from previous training steps is not accessible anymore, and we want to update the model to perform segmentation on new classes preserving its performance on old classes \ie $f_{\theta^{t}}:\set X\mapsto \real^{N\times |\set Y^{t}|}$.

\subsection{Training the Localizer} \label{sec:classifier}
Inspired by the WSSS literature~\cite{araslanov2020single, wang2020self, lee2021railroad, li2021pseudo, ahn2018learning, kolesnikov2016seed}, we introduce a \textit{localizer} $g$, trained with image-level labels, to produce the pseudo-supervision for the segmentation model.
The localizer uses the features from the segmentation encoder $e$ to predict a score for all classes (background, old and new ones) \ie $z = g(e(x)) \in \real^{|\set{Y}^t| \times H \times W}$.

\myparagraph{Learning from image-level labels.}
To learn from image-level labels first we need to aggregate the pixel-level classification scores $z$. The common solution is to use a Global Average Pooling (GAP) \cite{ahn2018learning, wang2020self}. 

{However, simply averaging the scores produces coarse pseudo-labels~\cite{araslanov2020single}, as all pixels in the feature map are encouraged to be less discriminative for the target class.}
For this reason, we use the \textit{normalized Global Weighted Pooling} (nGWP) \cite{araslanov2020single}, that weights every pixel based on its relevance for the target class.
In particular, the weight of each pixel is computed normalizing the classification scores with the \texttt{softmax} operation $\psi$, \ie $m = \psi(z)$. 
The aggregated scores are computed as:
\begin{equation} \label{eq:aggregation}
    \hat{y}^{nGWP} = \frac{\sum_{i \in \set I} m_i z_i}{\epsilon + \sum_{i \in \set I} m_i},
\end{equation}
where $\epsilon$ is a small constant. 
Moreover, to encourage the scores to identify all the visible parts of the object, we use the \textit{focal penalty} term introduced by~\cite{araslanov2020single}, that is obtained as:
\begin{equation}\label{eq:focal}
    \hat{y}^{FOC} = (1- \frac{\sum_{i \in \set I}m_i}{|\set I|})^\gamma log(\lambda + \frac{\sum_{i \in \set I}m_i}{|\set I|}),
 \end{equation}
where $\lambda$ and $\gamma$ are hyper-parameters. We refer the readers to~\cite{araslanov2020single} for more details on the nGWP and the focal penalty.

Since \wils\ is an incremental learning scenario, we assume to have access only to image-level annotations $y$ for the \textit{new classes} $\set{C}^t$. The localizer is then trained minimizing the \textit{multi-label soft-margin loss}:
\begin{multline}
\label{eq:loss_cls}
        \ell_{CLS}(\hat{y}, y) = - \frac{1}{|\set{K}|} \sum_{c \in \set{K}} y^c log(\hat{y}^c) + \\ + (1-y^c) log(1 - \hat{y}^c),
\end{multline} 
where $\set{K} = \set C^t$, $\hat{y} = \sigma(\hat{y}^{nGWP} + \hat{y}^{FOC})$, and $\sigma$ is the logistic function.
We note that, while the loss is computed only on new classes, it implicitly depends on the old classes scores due to the softmax-based aggregation in \cref{eq:aggregation}.  
However, since image-level annotations are cheap, and new images can be easily annotated, we may also consider a relaxed setting in which weak annotations are provide for both old and new classes. In this scenario the classification loss in~\cref{eq:loss_cls} is computed on all classes and $\set{K} = \set Y^t$.

\myparagraph{Localization Prior.}
The image-level labels provide supervision only on the presence of new classes in the image. However, they do not provide any cue on their boundaries or any information about the location of old classes.
We argue that these insights can be freely obtained from the segmentation model learned in previous learning steps. In particular, the background score can be used as a saliency prior to extracting better object boundaries. Moreover, the scores of the old classes guide the localizer in detecting whether and where an old class is present in the image, directing its attention to alternative regions.

Hence, we introduce a direct supervision on the localizer coming from the segmentation model trained on step $t-1$, \ie $f_\theta^{t-1}$. 
The supervision acts as a \textit{Localization Prior} (LOC) and can be provided as a pixel-wise loss between the segmentation model outputs $\omega = \sigma(f_\theta^{t-1}(x))$ and the classification scores $z$. Formally, we minimize the following objective function:
\begin{multline}\label{eq:loss_ocp}
    \ell_{LOC}(z, \omega) = - \frac{1}{|\set{Y}^{t-1}||\set I|} \sum_{i \in \set I} \sum_{c \in \set{Y}^{t-1}}  \omega^c_i log(\sigma(z^c_i)) + \\ + (1-\omega^c_i) log(1 - \sigma(z^c_i))),
\end{multline}
where $\sigma(\cdot)$ is the logistic function.

In \cref{eq:loss_ocp}, the segmentation model provides a dense target on old classes. Unlike the \texttt{softmax} operator, which enforces competition among classes, the \texttt{logistic} function makes the class probabilities independent which is beneficial for a correct localization prior; in the case of a novel class, both old classes and the background will have a low score, implicitly informing the localizer that the pixel belongs to a new class.

\subsection{Learning to Segment from Pseudo-Supervision} \label{sec:segmentation}

A solution often adopted by WSSS methods to train the semantic segmentation network is to extract hard-pseudo labels from an image-level classifier. In particular, these are obtained generating a one-hot distribution $q^{\text{Hard},c}$ for each pixel, attributing value one to the class with the maximum score for each pixel and zero to the others, \ie 
\begin{equation}
    q^{\text{Hard}, c}_i = \begin{cases}
        1 \;\;& \text{if}\ c=\argmax_{k\in{\set{Y}^t}} m^c_i,\\
        0 \;\;& \text{otherwise},
    \end{cases}
\end{equation}
where $m$ is the softmax normalized score extracted from the localizer. 

However, it is well-known that pseudo-supervision generated from an image-level classifier is noisy \cite{li2021pseudo, lee2021railroad, araslanov2020single, wang2020self}, and using $q^{H, c}$ to supervise the segmentation network may be detrimental for learning, causing the model to fit the wrong targets. 
For this reason, we propose to smooth the pseudo-labels to reduce the noise \cite{lukasik2020does}. Formally, given a class $c$, the pseudo-supervision $q^c$ is computed as:
\begin{equation}
\label{eq:smooth}
    q^c = \alpha q^{\text{Hard}, c} + (1-\alpha) m^{c},
\end{equation}
where $\alpha$ is a hyper-parameter that controls the smoothness.

Although the localizer produces scores for both new and old classes, the output distribution might be biased towards new classes due to the incremental training step. Thus, using $q$ as a target for the segmentation model would lead to catastrophic forgetting \cite{mccloskey1989catastrophic}.
Inspired by the knowledge distillation framework~\cite{hinton2015distilling}, we replace the pseudo-supervision extracted from the localizer on old classes with the output of the segmentation model trained in the previous learning step.
The final pixel-level pseudo-supervision $\hat{q}$ is thus composed as follows:
\begin{equation}
    \hat{q}^c = \begin{cases}
      \min(\sigma(f_{\theta^{t-1}}(x))^c, q^c) \;\;& \text{if}\ c=\mathtt{b},\\
      q^c \;\;& \text{if}\ c\in{\set{C}}^t, \\
      {\sigma(f_{\theta^{t-1}}(x))}^c \;\;& \text{otherwise},
    \end{cases}
\end{equation}
where $\mathtt{b}$ is the background class and $\sigma(\cdot)$ is the logistic function.
We note that we utilize the minimum value of the two distributions for the background class, which contributes in modeling the background shift \cite{cermelli2020modeling}.

Since the pseudo-supervision $\hat{q}^c$ is not a probability distribution that sums to one, as required by the standard softmax-based cross-entropy loss,
we propose to use a training loss based on the multi-label soft-margin loss: 
\begin{multline} \label{eq:loss_seg}
        \ell_{SEG}(p, \hat{q}) = - \frac{1}{|\set I|} \sum_{i \in \set I} \sum_{c \in \set{Y}^t} \hat{q}^c_i log(p^c_i) + \\ + (1-\hat{q}^c_i) log(1 - \sigma(p^c_i)),
\end{multline} 
where $\set{Y}^t$ is the set of all seen classes and $p=f_{\theta^t}(x)$ is the segmentation model output.

In conclusion, we remark that the localizer is not employed during the testing phase, thus our method does not increase the time required for the inference.

\section{Experiments}
\label{sec:experiments}
\subsection{Datasets and Settings}
We provide an extensive evaluation of \wilson\ on the two standard benchmarks Pascal VOC 2012~\cite{Everingham2009ThePV} and COCO~\cite{lin2014microsoft}. 
Following the standard methodology~\cite{kolesnikov2016seed, ahn2018learning}, we augment the Pascal VOC dataset with images from~\cite{hariharan2011semantic} for a total of $10582$ images for training and $1449$ for validation annotated on $20$ object categories.
COCO is a large-scale dataset providing 164K images and $80$ object classes. We follow the training split and the annotation of \cite{caesar2018coco} that solves the overlapping annotation problem present in \cite{lin2014microsoft}.

Following prior works~\cite{cermelli2020modeling, maracani2021recall}, we adopt two incremental learning settings on the Pascal VOC dataset: the \textbf{15-5 VOC}, where $15$ classes are learned in the first learning phase and $5$ new classes added in a second step, and the \textbf{10-10 VOC}, where two steps of $10$ classes are performed. Following \cite{cermelli2020modeling, maracani2021recall}, we report results using two experimental protocols: (i) the \textit{disjoint} scenario, in which each training step includes images containing only new or previously seen classes; (ii) the \textit{overlap} scenario, in which each training step includes all the images containing at least one pixel from a novel class.
In addition, we propose a novel incremental learning scenario, the \textbf{COCO-to-VOC}, composed of two training steps. First, we learn the $60$ COCO classes not present in the Pascal VOC dataset, removing all the images containing at least one pixel of the latter. Then, in the second step, we learn 20 Pascal VOC classes.
Following previous protocols \cite{cermelli2020modeling, maracani2021recall}, we report the results on the dataset validation sets since the test set labels have not been publicly released. We adopt the standard mean Intersection over Union metric (mIoU) \cite{Everingham2009ThePV} to evaluate the performance of the segmentation model.

We recall that, differently from \cite{cermelli2020modeling, maracani2021recall}, in the proposed \wils\ setting the incremental steps provide only image-level labels for the new classes.

\subsection{Baselines}
Given that {\wils} is a new setting, we compare \wilson\ with both incremental learning and weakly supervised semantic segmentation approaches.
We report eight methods that represent the current state-of-the-art for incremental learning using pixel-wise supervision: LWF~\cite{li2017learning}, LWF-MC~\cite{rebuffi2017icarl}, ILT~\cite{michieli2019incremental},  MiB~\cite{cermelli2020modeling}, PLOP~\cite{douillard2021plop}, CIL~\cite{klingner2020class}, SDR~\cite{michieli2021continual}, and RECALL~\cite{maracani2021recall}.
We note that RECALL~\cite{maracani2021recall}, differently from other methods, uses additional images taken from the Web.
For Pascal VOC, we use the results published in \cite{maracani2021recall, douillard2021plop}, while we run the experiments on the COCO-to-VOC setting using the code provided by \cite{cermelli2020modeling}.

Furthermore, we report the performance of several state-of-the-art WSSS methods adapted to operate in the incremental learning scenario. 
In particular, we first train a classification model using the images available in the incremental learning steps. Then, we generate the hard pseudo-labels offline and train the segmentation model minimizing~\cref{eq:loss_seg}.
We report the results with the pseudo-labels generated from: the class activation maps obtained from a standard image classifier (CAM), SEAM~\cite{wang2020self}, SS~\cite{araslanov2020single}, and EPS~\cite{lee2021railroad}. 
As for \wilson\, we followed the same experimental protocols provided by~\cite{cermelli2020modeling}, training each method using only the images belonging to disjoint and overlap scenarios. For each method, we used the implementation released by the authors to produce the results. For CAM, we used the implementation of EPS to generate the pseudo-labels. 
It is important to remark that, while CAM, SS, and SEAM rely only on image-level labels, EPS also makes use of an off-the-shelf saliency detector trained on external data.

\subsection{Implementation Details}
We employ Deeplab V3~\cite{chen2017deeplab} architecture for all the experiments, with a ResNet-101~\cite{he2016deep} backbone with output stride equal to $16$ for Pascal VOC and a Wide-ResNet-38~\cite{wu2019wider} with output stride $8$ for COCO, both pre-trained on ImageNet. As in \cite{cermelli2020modeling}, we use in-place activated batch normalization~\cite{rotabulo2017place} to reduce the memory footprint required by the experiments.
The localizer used to generate the CAMs is composed of 3 convolutional layers followed by batch normalization and Leaky ReLU, where the first two have kernel size $3\times3$ while the last $1\times1$, with channel numbers \{$256$, $256$, number of classes\}, and stride $1$.
The model is trained for $40$ epochs using batch size $24$ and SGD with an initial learning rate of $0.001$ ($0.01$ for the Deeplab head and the localizer), momentum $0.9$, and weight decay $10^{-4}$. We train only the localizer for the first $5$ epochs. Then, we train the whole network by adding the pseudo-supervision from the localizer and decay the learning rate using a polynomial schedule with a power of $0.9$.
Following \cite{araslanov2020single}, we set $\lambda=0.01$, $\gamma=3$ of~\cref{eq:focal}, and after the fifth epoch, we use the self-supervised segmentation loss on the localizer. Finally, we set $\alpha=0.5$ in~\cref{eq:smooth} for all the experiments. 
\begin{table}[t]
\centering

\begin{adjustbox}{width=1.0\columnwidth}
\vspace{-10pt}
\begin{tabular}{ll|ccc|ccc}
                &              & \multicolumn{3}{c|}{\textbf{Disjoint }}                                                                     & \multicolumn{3}{c}{\textbf{Overlap }}                                                                       \\
\textbf{Method} & \textbf{Sup} & \multicolumn{1}{l}{\textbf{1-15 }} & \multicolumn{1}{l}{\textbf{16-20}} & \multicolumn{1}{l|}{\textbf{All}} & \multicolumn{1}{l}{\textbf{1-15 }} & \multicolumn{1}{l}{\textbf{16-20}} & \multicolumn{1}{l}{\textbf{All}}  \\ 
\hline
Joint $\star$           & Pixel        & 75.5                               & 73.5                               & 75.4                              & 75.5                               & 73.5                               & 75.4                              \\ 
\hline
FT $\star$               & Pixel        & 8.4                                & 33.5                               & 14.4                              & 12.5                               & 36.9                               & 18.3                              \\
LWF $\star$ \cite{li2017learning}             & Pixel        & 39.7                               & 33.3                               & 38.2                              & 67.0                               & 41.8                               & 61.0                              \\
LWF-MC  $\star$ \cite{rebuffi2017icarl}               & Pixel        & 41.5                               & 25.4                               & 37.6                              & 59.8                               & 22.6                               & 51.0                              \\
ILT $\star$ \cite{michieli2019incremental}             & Pixel        & 31.5                               & 25.1                               & 30.0                              & 69.0                               & 46.4                               & 63.6                              \\
CIL $\star$ \cite{klingner2020class}                  & Pixel        & 42.6                               & 35.0                               & 40.8                              & 14.9                               & 37.3                               & 20.2                              \\
MIB $\star$ \cite{cermelli2020modeling}             & Pixel        & 71.8                               & 43.3                               & 64.7                              & 75.5                               & 49.4                               & 69.0                              \\
PLOP $\diamond$ \cite{douillard2021plop}          & Pixel        & 71.0                               & 42.8                               & 64.3                              & \underline{75.7}                              & 51.7                               & \underline{70.1}                              \\
SDR $\star$ \cite{michieli2021continual}              & Pixel        & \underline{73.5}                               & 47.3                               & \underline{67.2}                              & 75.4                               & 52.6                               & 69.9                              \\
RECALL $\star$ \cite{maracani2021recall}    & Pixel        & 69.2                               & \underline{52.9}                               & 66.3                              & 67.7                               & \underline{54.3}                               & 65.6                              \\ 
\hline
CAM             & Image        & 69.3	&	26.1	&	59.4	&	69.9	&	25.6	&	59.7	\\
SEAM \cite{wang2020self}           & Image        & 71.0	&	33.1	&	62.7	&	68.3	&	31.8	&	60.4	\\
SS  \cite{araslanov2020single}            & Image        & 71.6	&	26.0	&	61.5	&	72.2	&	27.5	&	62.1	\\
EPS  \cite{lee2021railroad}            & Image        & 72.4	&	38.5	&	65.2	&	69.4	&	34.5	&	62.1	\\ 
\hline
\textbf{\wilson\ (ours)}   & Image      &  \textbf{73.6}	&	\textbf{43.8}	&	\textbf{67.3}	&	\textbf{74.2}	&	\textbf{41.7}	&	\textbf{67.2}	\\

\end{tabular}

\end{adjustbox}
\caption{Results on the 15-5 setting of Pascal VOC expressed in mIoU\%. The best method using Image-level supervision is bold. The best method using Pixel supervision is underlined. $\star$: results from \cite{maracani2021recall}. $\diamond$: results from \cite{douillard2021plop}.
}
\label{table:voc_exp_15-5}
\end{table}
\begin{table}[t]
\centering

\begin{adjustbox}{width=1.0\columnwidth}
\vspace{-10pt}
\begin{tabular}{ll|ccc|ccc}
                     &              & \multicolumn{3}{c|}{\textbf{Disjoint }}                                                                     & \multicolumn{3}{c}{\textbf{Overlap }}                                                                       \\
\textbf{Method}      & \textbf{Sup} & \multicolumn{1}{l}{\textbf{1-10 }} & \multicolumn{1}{l}{\textbf{11-20}} & \multicolumn{1}{l|}{\textbf{All}} & \multicolumn{1}{l}{\textbf{1-10 }} & \multicolumn{1}{l}{\textbf{11-20}} & \multicolumn{1}{l}{\textbf{All}}  \\ 
\hline
Joint $\star$         & Pixel        & 76.6                               & 74.0                               & 75.4                              & 76.6                               & 74.0                               & \textit{75.4}                     \\ 
\hline
FT $\star$            & Pixel        & 7.7                                & 60.8                               & 33.0                              & 7.8                                & 58.9                               & 32.1                              \\
LWF $\star$ \cite{li2017learning}           & Pixel        & 63.1                               & 61.1                               & 62.2                              &  \underline{70.7}                               & 63.4                               & 67.2                              \\
LWF-MC  $\star$ \cite{rebuffi2017icarl}               & Pixel        & 52.4                               & 42.5                               & 47.7                              & 53.9                               & 43.0                               & 48.7                              \\
ILT $\star$ \cite{michieli2019incremental}           & Pixel        & \underline{67.7}                               &  \underline{61.3}                               &  \underline{64.7}                              & 70.3                               & 61.9                               & 66.3                              \\
CIL $\star$ \cite{klingner2020class}                  & Pixel        & 37.4                               & 60.6                               & 48.8                              & 38.4                               & 60.0                               & 48.7                              \\
MIB $\star$ \cite{cermelli2020modeling}          & Pixel        & 66.9                               & 57.5                               & 62.4                              & 70.4                               & 63.7                               & 67.2                              \\
PLOP \cite{douillard2021plop} & Pixel & 63.7	&	60.2	&	63.4	&	69.6	&	62.2	&	67.1	\\
SDR $\star$ \cite{michieli2021continual}           & Pixel        & 67.5                               & 57.9                               & 62.9                              & 70.5                               &  \underline{63.9}                               &  \underline{67.4}                              \\
RECALL $\star$ \cite{maracani2021recall}  & Pixel        & 64.1                               & 56.9                               & 61.9                              & 66.0                               & 58.8                               & 63.7                              \\ 
\hline
CAM                         & Image         & \textbf{65.4}	&	41.3	&	54.5	&	\textbf{70.8}	&	44.2	&	58.5	\\  
SEAM \cite{wang2020self}    & Image        & 65.1	&	53.5	&	60.6	&	67.5	&	55.4	&	62.7	\\
SS  \cite{araslanov2020single}  & Image    & 60.7	&	25.7	&	45.0	&	69.6	&	32.8	&	52.5	\\
EPS  \cite{lee2021railroad}  & Image       & 64.2	&	54.1	&	60.6	&	69.0	&	57.0	&	64.3	\\
\hline
\textbf{\wilson\ (ours)}     & Image       & 64.5	&	\textbf{54.3}	&	\textbf{60.8}	&	70.4	&	\textbf{57.1}	&	\textbf{65.0}	\\
\end{tabular}
\end{adjustbox}
\caption{Results on the 10-10 setting of Pascal VOC expressed in mIoU\%. The best method using Image-level supervision is bold. The best method using Pixel supervision is underlined. $\star$:results from \cite{maracani2021recall}.}
\label{table:voc_exp_10-10}
\vspace{-10pt}
\end{table}
\begin{table}[ht]
\centering

\setlength{\tabcolsep}{8pt} 

\begin{adjustbox}{width=1.0\columnwidth}
\begin{tabular}{ll|ccc|c}
                &              & \multicolumn{3}{c|}{{\textbf{COCO}}}                                                                   & {\textbf{VOC}}  \\
\textbf{Method} & \textbf{Sup} & \multicolumn{1}{c}{\textbf{1-60 }} & \multicolumn{1}{c}{\textbf{61-80 }} & \multicolumn{1}{c|}{\textbf{All}} & \textbf{61-80}         \\ 
\hline
FT                                    & Pixel        & 1.9	&	41.7	&	12.7	&	\underline{75.0} \\
LWF \cite{li2017learning}             & Pixel        & 36.7	&	\underline{49.0}	&	\underline{40.3}	&	73.6 \\
ILT \cite{michieli2019incremental}    & Pixel        & \underline{37.0}	&	43.9	&	39.3	&	68.7 \\
MIB \cite{cermelli2020modeling}       & Pixel        & 34.9	&	47.8	&	38.7	&	73.2 \\
PLOP \cite{douillard2021plop}         & Pixel        & 35.1	&	39.4	&	36.8	&	64.7 \\
\hline
CAM             & Image        &  30.7                                  &   20.3                                  &     28.1                              & 39.1                   \\
SEAM \cite{wang2020self}            & Image        &     31.2                               &       28.2                               &       30.5                            &       48.0                 \\
SS \cite{araslanov2020single}              & Image        & 35.1                               & 36.9                                & 35.5                              & 52.4                   \\
EPS \cite{lee2021railroad}            & Image        & 34.9                               & {38.4}                               & 35.8                              & 55.3                   \\ 
\hline
\textbf{\wilson\ (ours) }            & Image        & \textbf{39.8} &	\textbf{41.0} &	\textbf{40.6} &	\textbf{55.7} \\
\end{tabular}
\end{adjustbox}
\caption{Results on the COCO-to-VOC setting expressed in mIoU\%. The best method using Image-level supervision is bold.  The best method using Pixel supervision is underlined.}
\label{table:cocovoc_exps}
\vspace{-10pt}
\end{table}  %

\subsection{Results}
\myparagraph{Single step addition of five classes (15-5).}
In this setting, after the initial learning stage, the following $5$ classes of the VOC dataset are added: \textit{plant, sheep, sofa, train, tv-monitor}. We report results in~\cref{table:voc_exp_15-5}. 
Despite being trained only with image-level labels, \wilson\ achieves competitive results in all settings (disjoint and overlap) against approaches trained with pixel-wise supervision. Considering all the classes, in the disjoint scenario, we are able to outperform RECALL~\cite{maracani2021recall} by 1.0\% and SDR \cite{michieli2021continual} by 0.1\%, demonstrating the resilience of \wilson\ to forgetting without the need for a replay buffer while maintaining enough plasticity for learning new classes. Moreover, in the disjoint scenario, we surpass PLOP~\cite{douillard2021plop} by $1.0\%$ and MIB~\cite{cermelli2020modeling} by $0.5\%$ on new classes. 

Considering WSSS methods adapted to the \wils\ scenario, 
the results are a demonstration of the strengths of {\wilson}:
the ability to retain the knowledge of past classes and, most importantly, the capability of learning new semantic classes given only image-level annotations. Indeed, when considering new classes, we outperform EPS~\cite{lee2021railroad} by $+5.3\%$ mIoU in the disjoint scenario, although it uses saliency maps generated from an external off-the-shelf model. Moreover, SEAM~\cite{wang2020self} is outperformed by $11.7\%$ and SS~\cite{araslanov2020single} by $17.8\%$. 
These achievements are even more pronounced in the overlap scenario, where \wilson\ not only preserves all the prior knowledge but also achieves a $+7.2\%$ boost when learning new classes \wrt EPS. In this situation, the overall improvement is $+5.1\%$ when compared to the best methods (SS, EPS).  

\begin{figure*}[t]
    \centering
    \includegraphics[width=0.95\linewidth]{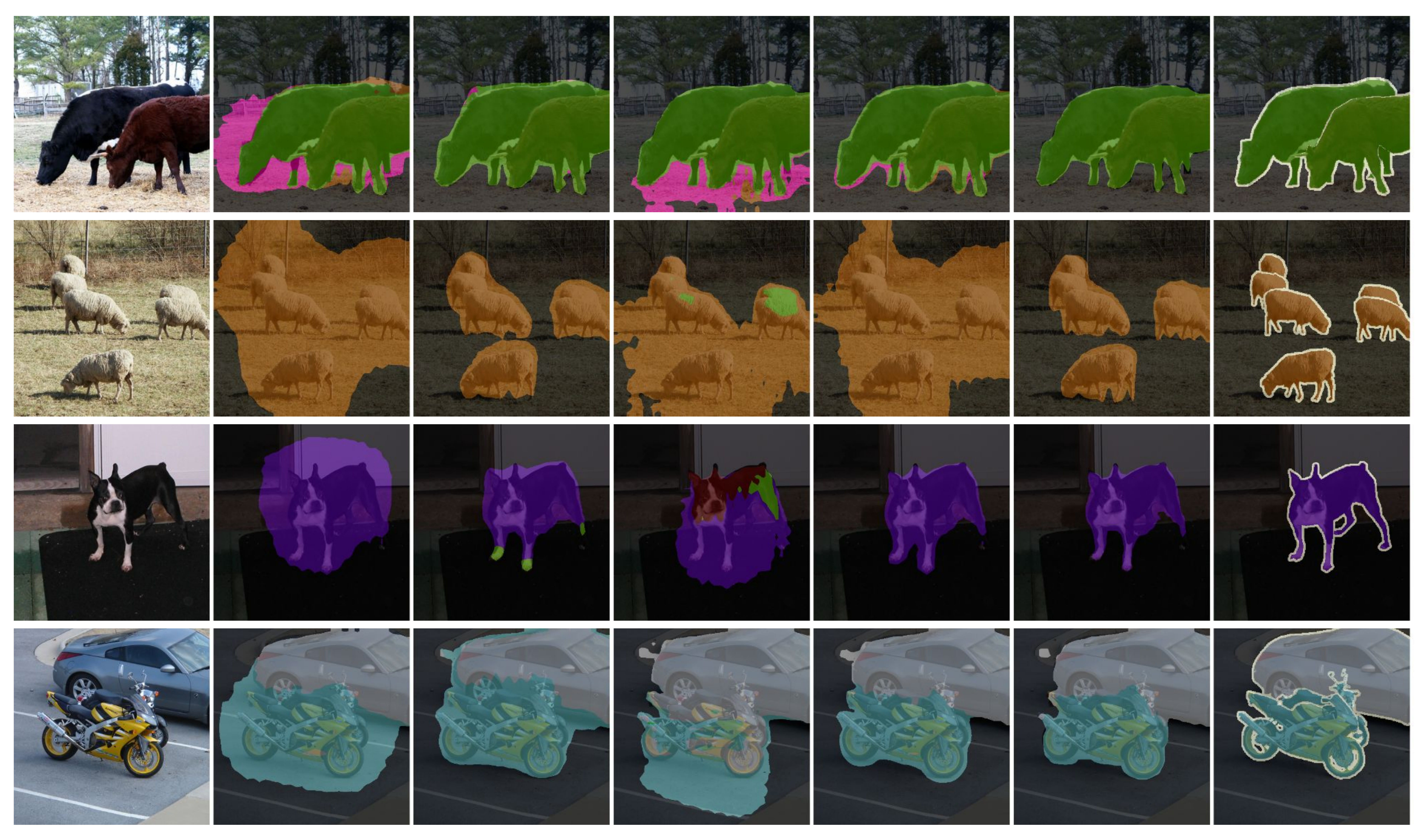}
    \vspace{-5pt}
     \caption{Qualitative results on the 10-10 VOC setting comparing different weakly supervised semantic segmentation methods. The image emphasized the efficiency of \wilson\ in both learning new classes (e.g. sheep, dog, motorbike) and preserving knowledge of old ones (e.g. cow, car). From left to right: image, CAM, SEAM \cite{wang2020self}, SS \cite{araslanov2020single}, EPS \cite{lee2021railroad}, \wilson\ and the ground-truth. Best viewed in color.}
    \vspace{-5pt}
    \label{fig:qualitative}
 \end{figure*}

\myparagraph{Single step addition of ten classes (10-10).} 
In this setting, we introduce $10$ classes in the incremental step: \textit{dining-table, dog, horse, motorbike, person, plant, sheep, sofa, train, tv-monitor}. \cref{table:voc_exp_10-10} shows consistent results with the 15-5 setting.
The differences between \wilson\ and IL (pixel-wise supervision) methods are quite small and the results are nearly comparable. In terms of accuracy, the gap using the most accurate incremental learning method, ILT, is $3.9\%$ in the disjoint scenario and shrinks to $2.4\%$ in the overlap one when compared to SDR.
The efficacy of \wilson\ is confirmed when compared to the WSSS (image-level supervision) method as well. Indeed, while learning novel semantic classes, our online technique outperforms all offline competitors in the overlap protocols by more than $+0.7\%$ overall mIoU, while achieving a comparable result (+0.2\%) in the disjoint scenario.
In Fig.~\ref{fig:qualitative} we report qualitative results demonstrating the superiority of \wilson\ on both new and old classes. 

\myparagraph{COCO-to-VOC.}
This set of experiments can be considered the most challenging. Initially, the network is trained on $60$ classes from the COCO dataset (which are not shared with VOC), while additional $20$ classes from the VOC dataset are added in the second step. 
\cref{table:cocovoc_exps} shows evaluations on both COCO and VOC validation sets. Despite the fact that \wilson\ performance drops 8\%
when learning new classes compared to LwF, this experiment better showcases our ability to retain prior information while learning new classes under image-level supervision, surpassing ILT performance on old classes (+2.8\%), which is the top competitor trained with pixel-wise supervision.
When comparing against WSSS methods, \wilson\ is the best method, marking $4.8\%$ improvements in terms of mIoU from the best WSSS method (EPS) on COCO. Similar results hold also for the VOC validation set. \wilson\ outperforms all the previous weakly supervised methods on both the old and new classes, both on COCO and VOC.

\subsection{Ablation studies} \label{sec:ablation}

\myparagraph{Localization Prior.}
To validate the robustness of the pseudo-supervision generation, we perform an ablation study considering different choices for training the localizer. Results are reported in \cref{table:ablation_bkg} on the VOC 10-10 disjoint and overlap scenarios. 
In particular, we compare different strategies for training the localizer: (i) we use a constant value for the old classes, as in \cite{araslanov2020single}, (ii) we use a fixed prior, directly concatenating the segmentation output of the old model to the class scores when computing $m$, (iii) we provide a localization supervision to the localizer with the {softmax} cross-entropy loss and (iv) with the loss in \cref{eq:loss_ocp}. 
Using a constant value and disregarding past knowledge from the old segmentation network results in lower performance when compared to the overall mIoU obtained if using a localization prior, particularly on new classes (-4.4\% on disjoint and -5.1 on overlap). This demonstrates that teaching the localizer the location of previous classes might be an effective way to prevent forgetting and improve performance while learning new classes. 
Thereby, using aggressive priors, such as directly using the segmentation output of the old model, does not allow the network to learn effectively the new classes, thus resulting in a gap of $-4.0\%$ on disjoint and $-4.3\%$ on overlapped scenario \wrt $\ell_{LOC}$.
Moreover, using the {softmax} cross-entropy loss to match the segmentation output is detrimental for the performance, achieving poor results on both new and old classes ($-6.3\%$ on disjoint and $-5.8\%$ on overlapped with respect to $\ell_{LOC}$). The reason for this result is that, due to the softmax normalization the cross-entropy loss does not consider each class independently, and forces the localizer to produce high scores for old classes even when they have low segmentation scores.

\begin{table}[t]
\centering

\begin{adjustbox}{width=1.0\columnwidth}

\begin{tabular}{cc|ccc|ccc}
                     && \multicolumn{3}{c|}{\textbf{Disjoint}}                             & \multicolumn{3}{c}{\textbf{Overlap}}            \\
\textbf{Prior} & \textbf{Loss} & \multicolumn{1}{l}{\textbf{1-10 }} & \textbf{11-20} & \textbf{All} & \textbf{1-10 } & \textbf{11-20} & \textbf{All}  \\ \hline
-       & - & 64.8                               & 49.9           & 58.8         & 69.4           & 52.0           & 62.0          \\
Fixed   & - & \textbf{66.1}                               & 50.3           & 59.7         & \textbf{71.4}           & 52.8           & 63.4          \\
Learned & CE  & 61.1                               & 46.0           & 54.5         & 67.6           & 49.5           & 59.2          \\
Learned & $\ell_{LOC}$ & 64.5	&	\textbf{54.3}	&	\textbf{60.8}	&	70.4	&	\textbf{57.1}	&	\textbf{65.0}	\\
\end{tabular}

\end{adjustbox}
\caption{Ablation study to validate the robustness of pseudo-supervision considering different types of localization priors for training the localizer.}
\label{table:ablation_bkg}
\end{table}
\begin{table}[t]
\centering

\setlength{\tabcolsep}{9pt} 

\begin{adjustbox}{width=1.0\columnwidth}

\begin{tabular}{l|ccc|ccc}
                      & \multicolumn{6}{c}{\textbf{VOC 15-5 }}                                                                                                   \\
\multicolumn{1}{c|}{} & \multicolumn{3}{c|}{\textbf{Disjoint }}                            & \multicolumn{3}{c}{\textbf{Overlap }}                               \\ 

\textbf{Method}       & \textbf{1-15 }       & \textbf{16-20}       & \textbf{All}         & \textbf{1-15}        & \textbf{16-20}       & \textbf{All}          \\
\hline
CAM                   & 70.5	&	34.7	&	62.6	&	71.6	&	36.0	&	63.7	\\
SEAM \cite{wang2020self} &71.9	&	26.9	&	61.7	&	70.8	&	28.1	&	61.0	\\
SS  \cite{araslanov2020single} &71.8	&	26.3	&	61.7	&	72.1	&	27.6	&	62.1	\\
EPS \cite{lee2021railroad} &73.5	&	45.7	&	67.7	&	75.3	&	\textbf{47.6}	&	69.4	\\
\textbf{\wilson\ (ours)}\ & \textbf{75.0}	&	\textbf{46.0}	&	\textbf{68.9}	&	\textbf{76.1}	&	45.6	&	\textbf{69.5}	\\

\multicolumn{1}{l}{}  & \multicolumn{1}{l}{} & \multicolumn{1}{l}{} & \multicolumn{1}{l}{} & \multicolumn{1}{l}{} & \multicolumn{1}{l}{} & \multicolumn{1}{l}{}  \\
                      & \multicolumn{6}{c}{\textbf{VOC 10-10 }}                                                                                                  \\
\multicolumn{1}{c|}{} & \multicolumn{3}{c|}{\textbf{Disjoint }}                            & \multicolumn{3}{c}{\textbf{Overlap }}                               \\
\multicolumn{1}{c|}{} & \textbf{1-10 }       & \textbf{11-20 }      & \textbf{All }        & \textbf{1-10 }       & \textbf{11-20 }      & \textbf{All }         \\ 
\hline

CAM                             & 63.1	&	42.2	&	53.9	&	66.6	&	45.0	&	56.8	\\
SEAM \cite{wang2020self}        & 66.0	&	50.4	&	59.7	&	70.9	&	54.6	&	64.0	\\
SS  \cite{araslanov2020single}  & 60.8	&	26.0	&	45.2	&	69.6	&	33.0	&	52.6	\\
EPS \cite{lee2021railroad}      & 69.1	&	53.0	&	62.4	&	72.9	&	55.7	&	65.4	\\
\textbf{\wilson\ (ours)}        & \textbf{69.5}	&	\textbf{56.4}	&	\textbf{64.2}	&	\textbf{73.6}	&	\textbf{57.6}	&	\textbf{66.7}	\\

\end{tabular}

\end{adjustbox}
\caption{Performance evaluation of weakly supervised segmentation methods trained with direct supervision on both old and new classes in the incremental step.}
\label{table:ablation_gt}
\vspace{-10pt}
\end{table}

\begin{figure}[t]
    \centering
    \includegraphics[width=\linewidth]{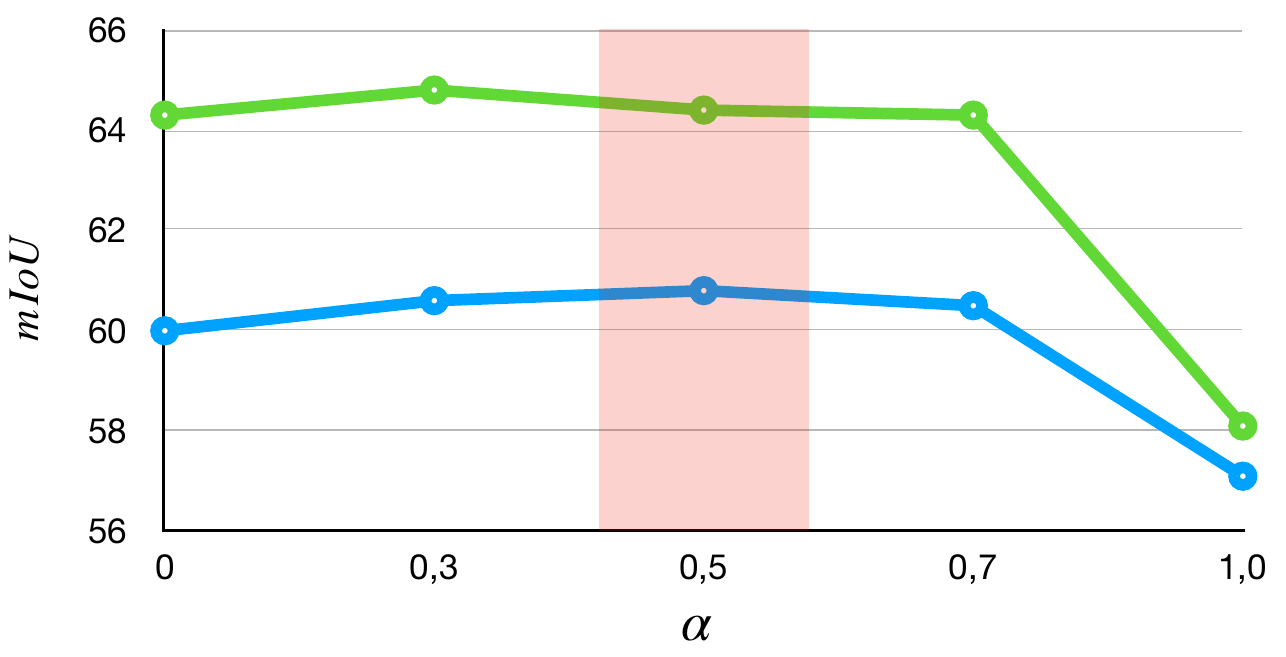}
    \caption{Ablation study about the effect of $\alpha$ to smooth the one-hot pseudo-labels used to supervise the $\ell_{SEG}$. Test reporting the mIoU for both the \textcolor{MidnightBlue}{Disjoint} and \textcolor{LimeGreen}{Overlap} VOC 10-10 protocols.}
    \vspace{-20pt}
    \label{fig:alpha_fig}
\end{figure}

\myparagraph{Smoothing effect on pseudo-supervision.}
We tune the hyper-parameter $\alpha$ of \cref{eq:loss_ocp}, which regulates the smoothness of the pseudo-labels supervising the segmentation model. In~\cref{fig:alpha_fig} we show the final mIoU in the VOC 10-10 disjoint and overlap scenarios, for five distinct $\alpha$ values ranging from $0$ to $1$. As expected, in the case of $\alpha=1$, which corresponds to using hard labels, the model fits the noise in the supervision, leading to worst results, forgetting the prior knowledge, and being incapable of learning novel classes. We chose $\alpha=0.5$ for our experiment since it is a reasonable trade-off in accuracy between learning and remembering. It is crucial to note that changing the values from $0$ to $0.7$ affects the results by less than $0.5\%$ on average between the disjoint and overlap case, indicating the robustness of \wilson\ to different $\alpha$ values.

\myparagraph{Using supervision for all the classes.}
Since image-level supervision is cheap, we evaluate the performance of weakly-supervised methods when the supervision is provided for both old and new classes in the incremental steps.
\cref{table:ablation_gt} reports the results on VOC. 
Comparing the results with \cref{table:voc_exp_15-5} and \cref{table:voc_exp_10-10}, we note a performance improvement. In particular, all the methods improved, with \wilson\ achieving, on average, 2\% on both old and new classes on the 15-5 and 10-10. This result demonstrates that introducing knowledge about old classes in the pseudo-supervision generation is crucial to both learning new classes and avoiding forgetting.
Moreover, we show that also in this scenario \wilson\ outperforms the offline WSSS methods. In particular, \wilson\ achieves better performance on every setting, outperforming EPS by 1.2\% and 0.1\% in the VOC 15-5 and by 1.8\% and 1.3\% in the VOC 10-10, respectively for the disjoint and overlapped scenario.

\subsection{Limitations}
\label{sec:limitations}
Despite the remarkable results achieved by \wilson, it still has some drawbacks. To begin with, it is unable to perform single-class incremental learning steps, since Eq.~\ref{eq:loss_cls} 
requires negative examples to properly guide the training. Moreover, we still need a considerable amount of images to train the model. Investigating learning from a few images could be an interesting future direction.

\section{Conclusions}
\label{sec:conclusions}
In this paper, we proposed \wils, a novel setting that aims to extend the knowledge of semantic segmentation models through cheap image-level annotations.
Applying current weakly-supervised learning approaches would require to generate the pseudo-supervision offline and then train the segmentation model.
Differently, we propose \wilson, that couples the semantic segmentation model with a localizer and use image-level annotations on the new classes to generate online the pseudo-supervision for the segmentation backbone. We show that adding a localization prior from the old model to the localizer improves the generation of the pseudo-labels. We prove the effectiveness of our approach in three incremental settings. We outperform the WSSS baselines that generate pseudo-labels offline and we get results close to fully supervised incremental learning methods.

{
\myparagraph{Acknowledgments.} We thank Francesco Visin, Carlo Masone, and Marco Cannici for their valuable comments on the early draft of the manuscript. We gratefully acknowledge the HPC infrastructure and Team at IIT.}

{\small
\bibliographystyle{ieee_fullname}
\bibliography{egbib}
}

\end{document}